# Similarity measure for aggregated fuzzy numbers from interval-valued data

Authors: Justin Kane Gunn[a], Hadi Akbarzadeh Khorshidi[b], Uwe Aickelin[c]
Email: justinkanegunn@gmail.com[a], hadi.khorshidi@unimelb.edu.au[b], uwe.aickelin@unimelb.edu.au[c]
School of Computing and Information Systems, The University of Melbourne, VIC 3010, Australia

**Abstract**

This paper presents a method to compute the degree of similarity between two aggregated fuzzy numbers from intervals using the Interval Agreement Approach (IAA). The similarity measure proposed within this study contains several features and attributes, of which are novel to aggregated fuzzy numbers. The attributes completely redefined or modified within this study include area, perimeter, centroids, quartiles and the agreement ratio. The recommended weighting for each feature has been learned using Principal Component Analysis (PCA). Furthermore, an illustrative example is provided to detail the application and potential future use of the similarity measure.

Keywords: Similarity Measure, Fuzzy Logic, Interval Agreement Approach, IAA, Interval-valued Data, Uncertainty.

## 1. Introduction

Similarity measures are a means quantifying the similarity between two data types, utilised by many modern algorithms. Areas covering algorithms that commonly require measurements of similarity within data include classification, ranking, decision-making and pattern-matching. A similarity measure can effectively substitute for a distance measure (e.g. Euclidean distance), making data types with defined similarity measures compatible with methods such as K-Nearest Neighbour [1, 2] and TOPSIS [3, 4, 5]. This study proposes a similarity measure for aggregate fuzzy numbers constructed from interval-valued data using the Interval Agreement Approach (IAA), that is when given two such fuzzy numbers the degree of similarity regarding them is computed.

The experimental interval-valued data in recent literature is often an alternative representation of expert opinion. Experts are commonly asked to summarise their opinions into a single numerical score, for example, "an 8 out of 10". This practice is seen throughout media reviewing (films, novels, music, etc), as well as for academic and job candidate selection. Scores are frequently aggregated and averaged to represent group opinion, then are used to rank the alternatives. However, as with all attempts to capture human response as data, the procedure of summarising opinions into a score consequently suffers from information loss.

The alternative mentioned that has received interest suggests recording interval-valued data as opposed to singular values. By interval-valued data, we are referring to data recorded as a range between two values. For example, a "6 to 8 out of 10". Interval scoring is one such method that allows experts to reflect the uncertainty of their judgement [6]. However, a challenge when using intervals in contrast to real numbers is that their operations are not as well defined; particularly operations on sets of intervals. Standard scores, being real numbers, can be aggregated, averaged and compared, which are universal data type operations required by many models that are not as clearly defined for interval-valued data. If such operations could be applied to interval-valued data, then in theory similar operations such as comparison and ranking could be calculated with increased effectiveness due to the preservation of uncertainty and reduced information loss. This had led to the development of the IAA method.

The IAA method, proposed by Wagner et al. [6], is a technique still in active research that retains all of the information provided by a given set of intervals and constructs them into a convenient data type known as a 'fuzzy number' [7]. A fuzzy number is specifically designed with the intention to capture and store uncertainty. We refer to aggregate fuzzy numbers from interval-valued data using IAA as IAA fuzzy numbers. IAA fuzzy numbers are reflective of the original sets of intervals of which they were constructed, thus the similarity measure proposed within this study is also effectively comparing sets of intervals. As IAA fuzzy numbers are relatively recent and were developed with the intent of flexibility, they are yet to have many of their operations agreed upon and fully defined, and as consequence applying IAA fuzzy numbers to various contemporary algorithms may prove challenging when a means of comparison between them is required. Intriguingly however, other strictly defined fuzzy number types, such as generalised fuzzy numbers, have had considerable amounts of research proposing useful applications of them, such as similarity and rankings measures [8, 9, 4, 10]. Through the abstraction and modification of operations previously defined for generalised fuzzy numbers, this study proposes a method for calculating the degree of similarity between two IAA fuzzy numbers. Developing a similarity measure for IAA fuzzy numbers allows for their further application into other areas of research, potentially as a solution to such areas that suffer from inaccuracies due to uncertainty within the data of which they model. When comparing sets of intervals that are derived from a group of expert opinions (typically through the use of a survey [6, 11, 12, 13]), the proposed similarity measure is computing the similarity of reception, and uncertainty thereof, between two sets of interval scores.

To propose the similarity measure, we first define new attributes for IAA fuzzy numbers, secondly we learn the weight of each attribute using an unsupervised machine learning method Principal Component Analysis (PCA) [14]. We then combine the attributes as features of comparison with the weights to produce the similarity measure, then finally we outline a synthetic dataset to illustrate its application and potential effectiveness.

## 2. Literature review

### 2.1. Interval-valued data represented by fuzzy numbers

The primary type of fuzzy number of which this research focuses on are those constructed from intervals, as an alternative to real numbers. Intervals, just being defined as a range between two real numbers, can be collected from many sources given various context. A frequent method of acquiring intervals throughout the recent literature is via survey, usually from a group of experts or critics in which intervals are explicitly requested. Alternatively, intervals are derived from words the experts or critics used to describe their opinion [15, 16, 12].

Liu and Mendel [16] first proposed a widely used method of converting a set of intervals to fuzzy numbers, known as the Interval Approach (IA). While the IA method was considered limiting regarding input data, it remained the only systematic method of constructing fuzzy numbers from a set of intervals for two years. Substantial modifications were then proposed by Coupland et al. [15], known as the Enhanced IA method (EIA), which was demonstrated to be an improvement over IA.

Miller et al. [17] built upon and discuss the findings of [15] and proposed a new method. Contrary to IA, the method [17] processes both intra-and inter-person viability, with intra-person referring to an expert's opinion changing over time and inter-person referring to a group of different expert opinions respectively. Miller et al. [17] also emphasise that their proposed method does not require prepossessing (stripping outliers, using thresholds etc), and thus no change or information is lost during its execution. After conducting tests on both synthetic and real-world data however, Miller et al. [17] concludes that both the proposed method, as well as IA have their separate advantages, and note that further comparative tests will have to be made.

Wagner et al. [6] merge their findings in [17] with IA, and as a result introduce a new major method of converting interval sets to Fuzzy Numbers, known as the Interval Agreement Approach (IAA). The IAA method offers a refined approach that includes the benefits of the method proposed in [17], and beyond this is shown to intentionally not make any assumptions (restrictions) about the input interval dataset. Thus, IAA produces a wide and unrestricted variety of fuzzy numbers relative to previous methods; which in theory fully represents and maintains the uncertainty that the intervals were initially intended to provide. Wagner et al. [6] conclude that IAA is a "highly useful method" for capturing interval-based (survey) data, stating that they now aim to drive further practical and theoretical developments of this method.

Recently, the IAA has seen increasing research interest. Potential improvements to IAA have been proposed, such as by Havens et al. [11], in which they outline another new method known as the efficient Interval Agreement Approach (eIAA). eIAA is shown to be more computationally efficient in contrast to the original IAA method, and beyond that provides a way to capture linguistical prototypes. Havens et al. [11] apply a case study involving film ratings as an example, but do emphasise that scenarios beyond expert opinions are quite feasible. Apart from proposed computational improvements, research involving IAA fuzzy numbers has resulted in the proposal of unique attributes. One such new attribute utilised throughout this study is known as the Agreement Ratio, proposed by Navarro et al. [12]. The Agreement Ratio is a proportional value between 0 and 1 that reflects the overall agreement amongst the experts given their survey responses. Along with other standard attributes, the Agreement Ratio has been shown to extract useful information and separate data adequately.

## 2.2. Similarity measures for generalised fuzzy numbers

Many different types of fuzzy sets and numbers are actively being researched and applied. One such type known as generalised fuzzy numbers have received much attention by researchers as they have shown to be useful in the area of decision-making, while maintaining a convenient and well-defined configuration. Boundaries such as the number of elements and their possible magnitudes have been defined for generalised fuzzy numbers, allowing researchers to produce elegant utilisations of them. Measures produced for generalised fuzzy numbers include ranking and similarity, making the area of study surrounding generalised fuzzy numbers an ideal reference for this research.

Chen [18] first proposed a similarity measure that given two generalised fuzzy numbers returns a value within the range of 0 and 1 that reflects their similarity. Ideally, a similarity measure would equate to 0 when the two input fuzzy numbers are entirely different, and 1 when they are precisely the same. However, due to the similarity measure [18] only utilising the geometric distance, it does not perform well on all occasions. Patra and Mondal [9] proposed a similarity measure that utilised geometric distance, area and height (highest degree of membership) of the input generalised fuzzy numbers. Khorshidi and Nikfalazar [8] proposed the similarity measure for generalised fuzzy numbers being one of the key inspirations for this research, which addresses these shortcomings of [9], primarily with the addition of the centroid (arithmetic mean) and perimeter attributes of generalised fuzzy numbers within the equation. It is through the abstraction and re-application of the similarity measure [8] for generalised fuzzy numbers that we have developed this study's proposed similarity measure for IAA fuzzy numbers.

## 3. Preliminaries

In this section, a brief description is introduced on the IAA method of constructing Type-1 Fuzzy Numbers from interval-valued datasets, and later a brief description of generalised fuzzy numbers and the similarity measure proposed in [8] that was the primary inspiration for this study.

### 3.1. IAA fuzzy numbers

$$A_1 = \{[1,2], [3,4], [3,6], [4,4]\}$$

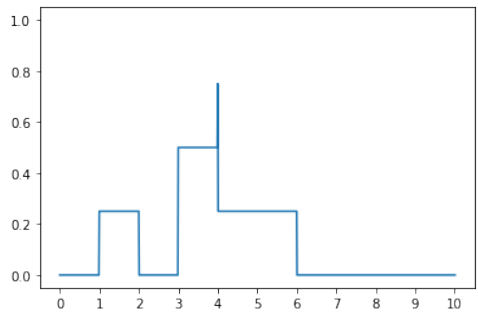

$$MC_{A_1} = [([1,2], 0.25), ((2,3), 0.0), ([3,4], 0.5), ([4,4], 0.7), ((4,6], 0.25)]$$

**Fig. 1.** Output of IAA given the set of intervals $A_1$, along with its $MC_A$ list.

A crisp interval is defined as $\bar{A} = [l_{\bar{A}}, r_{\bar{A}}]$, where $l$ is the 'left' start value and $r$ is 'right' end value, and thus a set of these intervals is defined as $A = \{\bar{A}_1, \dots, \bar{A}_n\}$. Given a set of intervals $A$ as input, the original IAA method [6, 11] is formulated as Eq (1). Where $y_i = i/n$, the degree of membership. Note that the use of the solidus " / " in Eq (1) is not referring to a division, but instead to the assignment of a given degree of membership to a set of values. The variable $y_i$ only equals 1 at values when all of the intervals of set $A$ overlap.

$$\mu_A = \sum_{i=1}^{n} {y_i} \Big/ \Big( \bigcup_{j_1=1}^{n-i+1} \bigcup_{j_2=j_1+1}^{n-i+2} \dots \bigcup_{j_i=j_{i-1}+1}^{n} \big(\bar{A}_{j_1} \cap \dots \cap \bar{A}_{j_i}\big) \Big) \qquad (1)$$

The original method, Eq (1), may not be intuitive and as such (in the context of Type-1 Fuzzy Numbers) it can be simplified into Eq (2, 3).

$$\mu_A(x) = \frac{\sum_{i=1}^{n} \mu_{\bar{A}_i}(x)}{n} \quad (2)$$

$$\mu_{\bar{A}_i}(x) = \begin{cases} 1 & l_{\bar{A}_i} \leq x \leq r_{\bar{A}_i}, \\ 0 & else. \end{cases} \quad (3)$$

In written terms, the IAA method defines the degree of membership of a variable real number $x$ for a given set of crisp intervals $A$, as the frequency (or count) of which $x$ appears within each interval divided by n (i.e. the cardinality of set $A$ / $|A|$ / the amount of intervals). Refer to Fig. 1 for example output of IAA given an elementary set of intervals.

### 3.2. Generalised fuzzy numbers

A fuzzy number in the traditional sense is any set of real numbers that also includes a degree of membership for each element (i.e. the magnitude of which each element belongs to said set), whilst a generalised fuzzy number is a more specified subclass. A generalised fuzzy number is one such type of fuzzy number represented by the following notation: $\tilde{A} = (a_1, a_2, a_3, a_4; w_{\tilde{A}})$. Where $a_{1-4}$ are the elements of the fuzzy number $\tilde{A}$, and $w_{\tilde{A}}$ is considered the height (maximum degree of membership) of the elements of said fuzzy number $\tilde{A}$. Depending on the relationship of the elements $a_{1-4}$ (the shape of the membership function curve produced) further subclasses of generalised fuzzy numbers are defined; for example, triangular, trapezoidal, crisp etc. Generalised fuzzy numbers having such requirements beyond that of traditional fuzzy numbers allows for their elegant application, particularly in areas such as risk analysis and decision making [8, 9, 4].

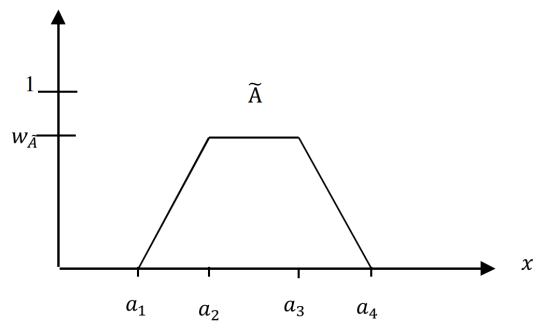

**Fig. 2.** Abstract representation of a generalised trapezoidal fuzzy number as seen in [8].

As discussed in the previous section, Khorshidi and Nikfalazar [8] proposed a group of useful attribute definitions that are then utilised to produce a similarity measure for two generalised fuzzy numbers. Excluding the variables to define the generalised fuzzy numbers themselves, the attributes described within [8] that produce the proposed similarity measure are area, perimeter, centroid-x and centroid-y, which follow their standard mathematical definitions given the curve produced by the membership function is defined as a geometric shape. Fig. 2 illustrates how the membership function over the x-y axis produces geometric shapes; in this case it produces a trapezoid. Given two generalised fuzzy numbers $\tilde{A}$ and $\tilde{B}$, the proposed similarity measure by [8] is defined by Eq (4). Where $A(\tilde{A})$ is the area of $\tilde{A}$, $P(\tilde{A})$ is the perimeter of $\tilde{A}$, and $d'(\tilde{A}, \tilde{B})$ is the centroid distance between $\tilde{A}$ and $\tilde{B}$ as defined by Eq (5).

$$S(\tilde{A}, \tilde{B}) = \left(1 - \frac{\sum_{i=1}^{4}|a_i - b_i|}{4} \times d'(\tilde{A}, \tilde{B})\right) \times \left(1 - \frac{\left(|A(\tilde{A}) - A(\tilde{B})| + |w_{\tilde{A}} - w_{\tilde{B}}| + \frac{|P(\tilde{A}) - P(\tilde{B})|}{\max\left(P(\tilde{A}), P(\tilde{B})\right)}\right)}{3}\right)$$

(4)

$$d'(\tilde{A}, \tilde{B}) = \frac{\sqrt{(Cx_{\tilde{A}} - Cx_{\tilde{B}})^2 + (Cy_{\tilde{A}} - Cy_{\tilde{B}})^2}}{\sqrt{1.25}}$$

(5)

In-depth theory, exploration and proofs as to the final construction of this similarity measure are depicted by Khorshidi and Nikfalazar in [8], they conclude that the similarity measure effectively handles the drawbacks of those proposed in previous works.

## 4. Proposed attributes

In this section, a description of all the attributes of IAA fuzzy numbers necessary for the proposed similarity measure are provided. Each algorithm throughout this section relies on the concept that IAA fuzzy numbers are representable as a list of tuples describing the regions ($R$) of change over the membership function $\mu_A(x)$, this list is referred throughout this study as Membership Curve ($MC_A$). Refer to Eq (6) and Fig. 1 for an abstract and practical illustration of the $MC_A$ list respectively. Where $R$ refers to a region of the curve, with left point, right point and height data. $MC_A$ is a list comprised of regions $R$ that reflect the membership function over the domain.

$$R_i = \left([R_{i_l}, R_{i_r}], R_{i_h}\right)$$
$$MC_A = [R_1, R_2, \ldots, R_{n-1}, R_n]$$

(6)

Where $l$ = left (start) point, $r$ = right (end) point, and $h$ = height (degree of membership)

### 4.1 Area

The area of an IAA fuzzy number, Eq (7), is the geometric area of its membership function curve. Currently, the definition of area has not changed from the geometric standard. Therefore, it is simply defined as the sum of the area of regions within the membership function curve. Given that all regions can be described as either rectangles or lines (i.e. rectangles of with an area of zero), the area of each region is defined as its width multiplied by its height.

$$A(MC_A) = \sum_{i=1}^{n} R_{i_h} \times (R_{i_r} - R_{i_l})$$

(7)

### 4.2 Perimeter

The perimeter of an IAA fuzzy number, Alg (1), the geometric perimeter of its membership function curve. That is, the perimeter is the sum of the edges (also referred to as margin, or boundary) produced by the shape of the curve. Perimeter both quantifies certainty and uncertainty as well as be responsive to small changes in the membership function curve.

| Algorithm 1 - Perimeter |
|---|
| ```
set perimeter = 0
for each R in MC_A
    if final R
        perimeter = perimeter + R_h
    else
        set width = R_r - R_l
    if width == 0
        //region is a vertical line
        set R_overlap = max(previous_R_h, R_h)
        perimeter = perimeter + |R_h - R_overlap|
    else
        perimeter = perimeter + width
        if R_h > 0
            //region is a rectangle
            perimeter = perimeter + width
            if first region
                previous_R_h defaults to 0
            perimeter = perimeter + |previous_R_h - R_h|
return perimeter
``` |

### 4.3 Centroid

The centroid, Eq (8 - 10), also known as the 'centre of mass' or 'weighted mean', is the arithmetic mean position of all points within a shape. To define the exact centre point of an entire shape, the location of the centroid on all axes must be given, so in the case of Type-1 Fuzzy Numbers there exists a centroid-x and centroid-y. The generic geometric centroid is calculated as the centre of each region of a shape, multiplied by the area of that region, the sum of which is then divided by the total area of the shape. In doing this it then 'weighs' each centre point, thus skewing the final overall centre point in the direction with the most area (mass).

The centroid has been modified to be susceptible to changes in both regions with and without area equally. To achieve this, the 'mass' has been redefined from geometric area to the degree of membership/height of each region (in contrast to the width multiplied by height) and each region as the start and end lines to each rectangle (for a single line that is also its location on the x-axis doubled). This variant method effectively considers the mass of both uncertain and certain areas equally.

$$centroid_x(MC_A) = \frac{\sum_{i=1}^{n}(R_{i_h} \times R_{i_l}) + (R_{i_h} \times R_{i_r})}{\sum_{i=1}^{n} 2(R_{i_h})} \quad (8)$$

$$centroid_y(MC_A) = \frac{\sum_{i=1}^{n} R_{i_h}/2}{\sum_{i=1}^{n} non\_zero(R_{i_h})} \quad (9)$$

$$non\_zero(x) = \begin{cases} 1 & x > 0, \\ 0 & else. \end{cases} \quad (10)$$

### 4.4 Quartiles

The quartiles are those defined in classical statistics. Three values that define various median data points of a given set of values. The first quartile being the median between the minimum value and the median of the set, the second quartile being simply the median of the set, and the third quartile being the median between the median of the set and the maximum value.

Regarding IAA fuzzy numbers, the quartiles attribute is a tuple of five values (also known as a five-number summary), the minimum, quartile 1, quartile 2, quartile 3 and maximum of the ordered set of left and right points, $l_{\bar{A}}$

and $r_{\bar{A}}$. The left and right points can either be obtained via the original interval set $A$, or extracted from the membership function $\mu_A(x)$ curve by normalizing the points of change on the x-axis. Either method yields the same results, the set of left and right points along with their respective frequencies can be separated into percentiles, providing the quartile data.

**4.5 Agreement ratio**

The agreement ratio, Alg (2), proposed by Navarro et al. [12] is a measure within the range of 0 and 1 of the agreement amongst the data sources (e.g. experts) that have provided their opinion as intervals. It has shown to be a useful unique statistic of IAA fuzzy numbers that quantify the overall group certainty of data sources. The agreement ratio is the only attribute utilised within this study that requires knowledge of the original set of intervals $A$ together with the respective IAA fuzzy number. The original definition of the agreement ratio [12] remains undefined for an IAA fuzzy number produced from a single interval (i.e. when $|A| = 1$), nor does it account for intervals that have equal left and right values (i.e. when $l_{\bar{A}} = r_{\bar{A}}$). As this study does account for these previously undefined sets of intervals, a proposal of some minor additions to the original definition of the agreement ratio is required.

Regarding the first issue of single interval sets, the agreement ratio of such a set is now defined to be consistently 1. The reason being is that from a purely logical perspective, it would be reasonable to assume that a single entry would be in complete agreement with itself. From a mathematical perspective, a definition of 1 is also fitting. The original algorithm utilises the degree of membership as a means of weighting the data, and an IAA fuzzy number constructed from a single interval only has one degree of membership that is 1. With no opposing intervals, it reasonably follows that the agreement ratio is 1.

Concerning the second issue of equal left and right values, they are usually included without error. However, they do produce single line regions and as consequence have a distance of zero. If a division happens to include a total distance of zero in the denominator, a division by zero error occurs. The agreement ratio algorithm now excludes iterations that would induce a division by zero error.

Note that intervals are not considered in agreement when a right point of one is equal to the left point of another, for example, the agreement ratio of set {[1,2], [2,3]} would evaluate to 0. Only when a right point of one interval is greater than the left point of another are they considered to share agreement, for example, agreement ratio of set {[1,2], [1.9,3]} would evaluate to 0.05.

For the agreement ratio to be computed, it requires 'alpha length' defined in Alg (3). Alpha length when given a real number $a$ and $MC_A$, computes the total length of all regions with a degree of membership equal to or above $a$.

Algorithm 2 – Agreement Ratio
```
set n = |A|
if n == 1
    return 1

set total = 0
set weights = 0
for i = n to 1
    if alphaLength(i_−1/n) != 0
        total = total + ((i/n) × (alphaLength(i/n) / alphaLength(i_−1/n)))
    weights = weights + (i/n)

return (total / weights)
```

Algorithm 3 – Alpha Length
```
set length = 0
for each region in MC_A
    if region_height ≥ a
        length = length + (region_end − region_start)

return length
```

## 5. Proposed similarity measure

In this section, a description of the proposed similarity measure is provided. The similarity measure proposed for IAA fuzzy numbers builds upon the theory of Eq (4), that is the similarity measure for generalised fuzzy numbers proposed by Khorshidi and Nikfalazar [8]. The proposed similarity measure of this study is a linear model that returns a value between 0 and 1 reflecting the similarity of two IAA fuzzy numbers. The features of this model are themselves similarity measures of the attributes described in the previous section, all in turn also returning a value between 0 and 1, with their weighted mean being the final output.

The weights were calculated using the Principal Component Analysis (PCA) method, an unsupervised Machine Learning technique useful for dimension reduction. PCA compresses the information of the features into lower numbers of components. The first principal component keeps the largest information of the dataset. The first principal component works as a weighted average where weights are factor loadings. So, we can come up with our linear model. The first principal component has been used for developing composite indices [14]. To learn the weights, we randomly generated a large dataset of IAA fuzzy numbers which were produced by interval sets. The absolute values of the weights outlined are an indication of the effectiveness each attribute had when separating the data under PCA, with their sum of squared values equalling 1. As is the consequence with random datasets the computations will fluctuate with every execution of the evaluation software, however, the weights proposed within this study were those that reflect the point of convergence, that is, a re-execution will result in weights of little difference. Perimeter, quartile, height and agreement attributes were all evaluated to have similar weights with little fluctuation in rank, with only area and centroid-x being substantially lower; implying that area and centroid-x are relatively not diverse amongst the data. The weights are by no means absolute and are expected to be modified given the context of varying scenarios. For example, quartiles are potentially more relevant when ranking IAA fuzzy numbers as they reflect direction and magnitude to ideal values, whilst perimeter and area are potentially more relevant when pattern-matching as they fluctuate due to changes in shape and not position.

The proposed similarity measure is outlined in Eq (11) and Table. 1. Note that $w_{\tilde{A}}$ refers to the height of $\tilde{A}$, that is the maximum degree of membership recorded across the domain, and the variable $a_i$ refers to the i-th quartile (five-number summary) of $\tilde{A}$.

$$S(\tilde{A}, \tilde{B}) = \left(1 - \sum_{i=1}^{6} w_i^2 f_i\right) \quad (11)$$

| | Feature vector - $f$ | Weight vector - $w$ |
|---|---|---|
| Quartile Distance: | $\dfrac{\sum_{i=1}^{5}|a_i - b_i|}{5(range)}$ (12) | 0.320726 |
| Centroid Distance: | $\dfrac{\sqrt{(Cx(\tilde{A}) - Cx(\tilde{B}))^2 + (Cy(\tilde{A}) - Cy(\tilde{B}))^2}}{\sqrt{(range)^2 + 0.5^2}}$ (13) | -0.509757 |
| Area Difference: | $\dfrac{|A(\tilde{A}) - A(\tilde{B})|}{\max(A(\tilde{A}), A(\tilde{B}))}$ (14) | 0.100985 |
| Height Difference: | $|w_{\tilde{A}} - w_{\tilde{B}}|$ (15) | -0.461649 |
| Perimeter Difference: | $\dfrac{|P(\tilde{A}) - P(\tilde{B})|}{\max(P(\tilde{A}), P(\tilde{B}))}$ (16) | 0.444451 |
| Agreement Ratio Difference: | $|AR(\tilde{A}) - AR(\tilde{B})|$ (17) | -0.465218 |

Table. 1. Similarity measure feature, Eq (12 - 17), and weight vectors.

When a global range is known: $range = maximum\ potential\ value - minimum\ potential\ value$

When a global range is unknown: $range = max(a, b) - min(a, b)$

A collection of diverse example uses of the proposed similarity measure are illustrated within Fig. 3. The global range was considered unknown and therefore used a local range for each example. Set 1 through to Set 6 are all demonstrations of ways in which differing uncertainty within the data, overlap of membership function $\mu_A(x)$ curves and height variations affect the evaluated similarity.

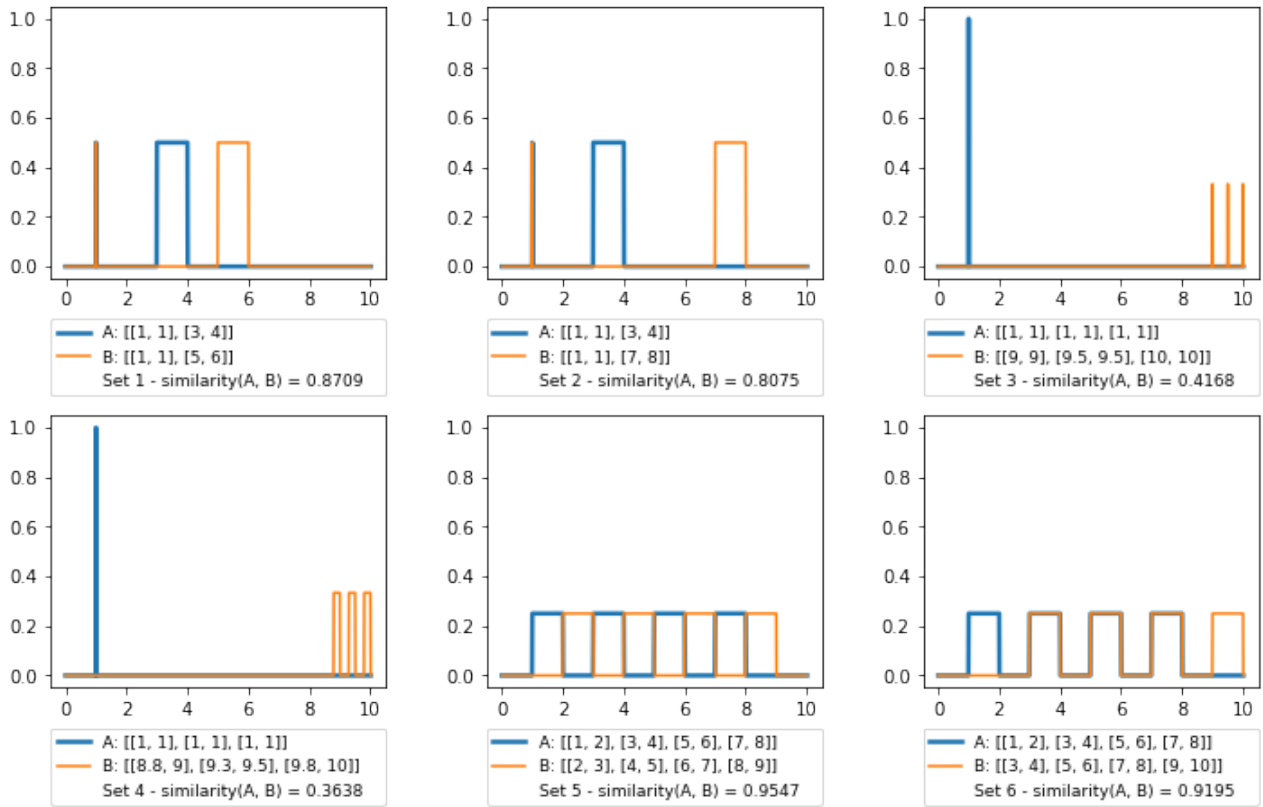

**Fig. 3.** Examples of the proposed similarity measure output on various IAA fuzzy numbers.

## 6. Illustrative example

In this section, an illustrative example study is provided to demonstrate and analyse the proposed measure described throughout this study. The dataset is synthetic, it has been provided to illustrate essential use, and emphasise specific benefits and issues of the implementation. Furthermore, the synthetic dataset is not confidential, and therefore open to those seeking to re-evaluate this experiment, or as a contrast when applying implementation in future research.

The following dataset, Table. 2, regards ten films that have been reviewed by five critics. Hypothetically, the critics were asked to give their opinion as an interval, of which crisp (i.e. when $l_{\bar{A}} = r_{\bar{A}}$) and decimals values were permitted, with scores being from 1 to 10.

|        | Critic 1  | Critic 2 | Critic 3   | Critic 4 | Critic 5 |
|--------|-----------|----------|------------|----------|----------|
| Film A | [1, 1]    | [1, 1]   | [1, 1]     | [1, 1]   | [1, 1]   |
| Film B | [5, 6]    | [6, 7]   | [10, 10]   | [3, 4]   | [5, 5]   |
| Film C | [2, 3]    | [1, 3]   | [4, 7]     | [1, 3]   | [4, 5]   |
| Film D | [6, 6]    | [6, 10]  | [8, 10]    | [5, 9]   | [2, 3]   |
| Film E | [1, 4]    | [2, 3]   | [7, 8]     | [3, 3]   | [2, 4.4] |
| Film F | [7, 7]    | [8, 9.2] | [9, 10]    | [8, 9]   | [9, 10]  |
| Film G | [8, 9]    | [9, 10]  | [9.5, 9.5] | [9, 10]  | [10, 10] |
| Film H | [1.5, 6.5]| [3, 10]  | [1, 10]    | [2, 9.3] | [8, 8.8] |
| Film I | [8, 8]    | [8, 8]   | [8, 8]     | [8, 8]   | [8, 8]   |
| Film J | [10, 10]  | [10, 10] | [10, 10]   | [10, 10] | [10, 10] |

**Table. 2.** Synthetic film review dataset

The interval-valued dataset has been intentionally constructed to demonstrate a wide range of possible IAA fuzzy numbers, illustrating use of the similarity measure. Each entry regards varying plausible reactions that groups may have towards a film. Film A is the worst possible response a group in this scenario may have, while in contrast, Film J is the ideal best response. Film B received a favourable, higher than average response. Film C received a mediocre, lower than average response. Film D received a favourable response, however, was disliked by Critic 5. Film E received a mediocre response; however, Critic 3 held it in high regard. Film F received a very favourable response. Film G received an extremely favourable response, which would likely be considered the highest 'realistic' response a film would receive in the real-world. Film H received an erratic response with no clear skew. Finally Film I received a crisp set of 8/10 scores, and while unlikely to occur in the real-world, nevertheless provides a further analysis of IAA fuzzy numbers bound to a specific crisp point.

Regarding the Critic selection. Critic 3 reflects an individual that generally responds well to films, and Critic 2 and Critic 4 were intentionally produced to have similar opinions. Each of the film's scoring set of intervals is then constructed into an IAA fuzzy number, as illustrated in Fig. 4. In the case of this illustrative example, the process required for comparing each set of intervals (i.e. aggregated reviews of each Film A to J) using the proposed similarity measure is outlined by the following steps:

1. Using either Eq (1) or Eq (2, 3) construct an IAA fuzzy number for every film using their respective sets of interval scores, with each interval set being a row in Table. 2. The output of such is given in Fig. 4.

2. Compute all attributes outlined in Section 4 for each IAA fuzzy number constructed in Step 1. Again, the output for such is given in Fig. 4.

3. Having computed each IAA fuzzy number along with their respective attributes, they are now comparable whilst being input for the proposed similarity measure described in Section 5; Eq (11). The output of the proposed similarity measure across all IAA fuzzy numbers constructed from Step 1 along with their attributes from Step 2 is given in Table. 3.

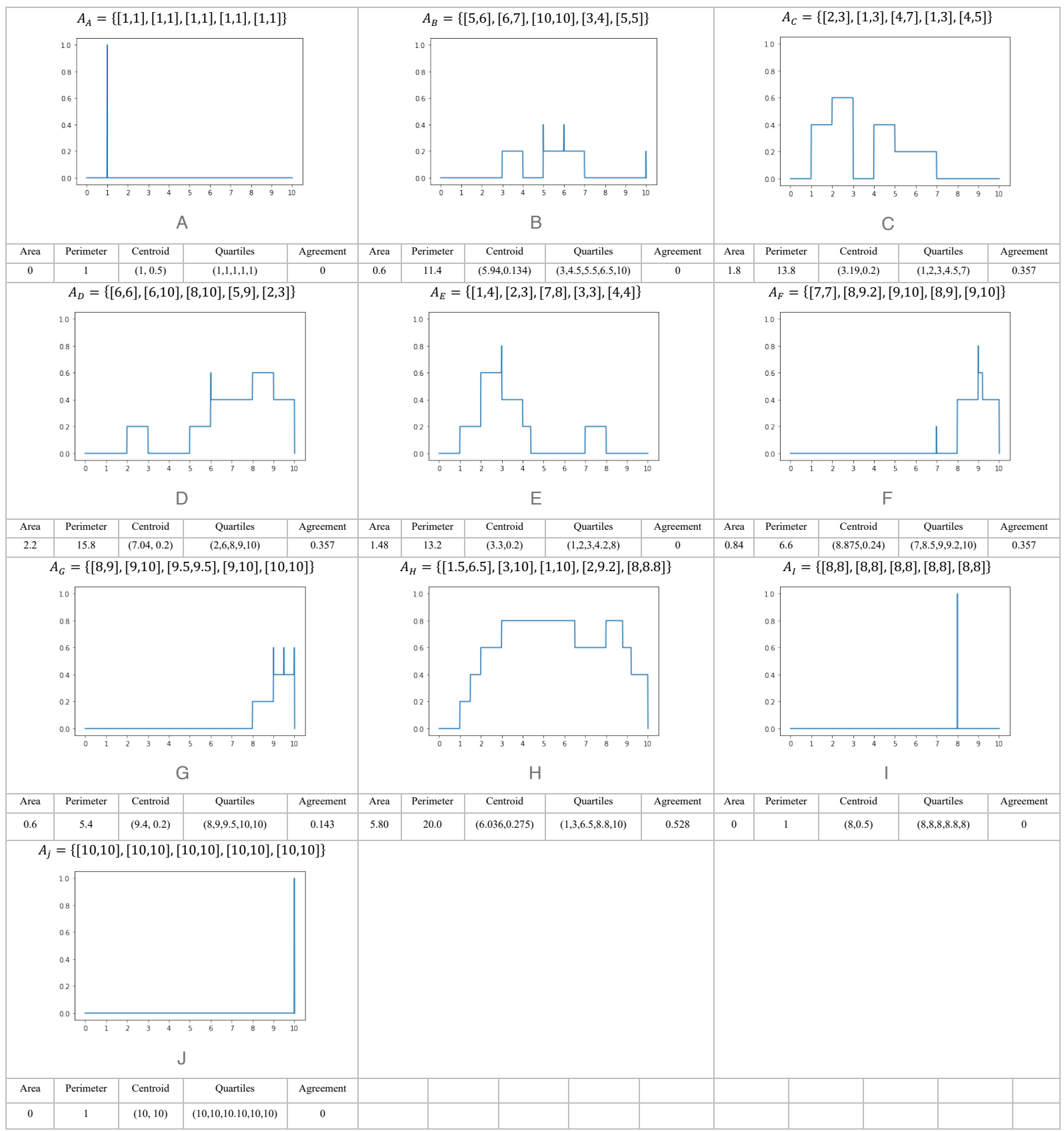

Fig. 4. IAA fuzzy number representation of the film dataset.

The proposed similarity measure is used on the synthetic dataset previously outlined. As the similarity measure was developed as a means of comparing IAA membership function curves, it is most useful when attempting to extract potentially important patterns within data. Without modification, the proposed similarity measure's primary functionality in this scenario would be to simply evaluate the similarity of reception to between each film. Note that the global range has now been defined within the constraints of this scenario, being evaluated to 9 (i.e. 10 - 1). The two separate films that are evaluated to have the highest degree of similarity are Film C and Film E, having a clearly similar structure over their membership function curves, that is two primary regions with the left half side being higher than the right. The two separate films evaluated to have the second highest degree of similarity are Film I and Film J, this is appropriate given their membership function curves are of the exact same shape with the only difference being their location on the x-axis. The two separate films that are evaluated to have the lowest degree of similarity are Film C and Film J, this is the result of Film C not only being very different in shape to Film I, but it is also placed within the lower half of the x-axis.

|        | Film A | Film B | Film C | Film D | Film E | Film F | Film G | Film H | Film I | Film J |
|--------|--------|--------|--------|--------|--------|--------|--------|--------|--------|--------|
| Film A | 1.0000 | 0.4830 | 0.5527 | 0.3993 | 0.5900 | 0.3867 | 0.3747 | 0.4444 | 0.7182 | 0.6377 |
| Film B | 0.4830 | 1.0000 | 0.7422 | 0.7686 | 0.7028 | 0.6342 | 0.6829 | 0.6859 | 0.5882 | 0.5173 |
| Film C | 0.5527 | 0.7422 | 1.0000 | 0.8120 | 0.9461 | 0.6308 | 0.5851 | 0.7351 | 0.4545 | 0.3740 |
| Film D | 0.3993 | 0.7686 | 0.8120 | 1.0000 | 0.7755 | 0.7633 | 0.7211 | 0.8305 | 0.5881 | 0.5222 |
| Film E | 0.5900 | 0.7028 | 0.9461 | 0.7755 | 1.0000 | 0.6781 | 0.5471 | 0.7843 | 0.5020 | 0.4215 |
| Film F | 0.3867 | 0.6342 | 0.6308 | 0.7633 | 0.6781 | 1.0000 | 0.8498 | 0.7072 | 0.6629 | 0.6546 |
| Film G | 0.3747 | 0.6829 | 0.5851 | 0.7211 | 0.5471 | 0.8498 | 1.0000 | 0.5836 | 0.6559 | 0.6865 |
| Film H | 0.4444 | 0.6859 | 0.7351 | 0.8305 | 0.7843 | 0.7072 | 0.5836 | 1.0000 | 0.5511 | 0.4835 |
| Film I | 0.7182 | 0.5882 | 0.4545 | 0.5881 | 0.5020 | 0.6629 | 0.6559 | 0.5511 | 1.0000 | 0.9195 |
| Film J | 0.6377 | 0.5173 | 0.3740 | 0.5222 | 0.4215 | 0.6546 | 0.6865 | 0.4835 | 0.9195 | 1.0000 |

**Table. 3.** Similarity amongst the film IAA fuzzy numbers outlined in Fig. 4.

For the purposes of this illustration, other intriguing information may be the similarity of opinions amongst the critics. When recommending films to a critic, in contrast to recommending the highest-rated film, offering them films based on the films that others with similar tastes to them have enjoyed (and conversely, avoiding films that they haven't enjoyed) may be more relevant to the critic personally.

A simple method of comparing the similarity between critics would be first to construct IAA fuzzy numbers out of their survey responses, then record the similarity of said fuzzy numbers. The sets of intervals regarding each critic is the transpose of the original dataset (matrix) Table. 2. Thus, the fuzzy numbers for each critic, along with their respective similarity measure outputs have been computed and contrasted, as seen in Fig. 5. As previously mentioned, Critic 2 and Critic 4 were intentionally synthetically produced with similar opinions in mind, and the proposed similarity measure does indeed successfully capture this as they are two separate critics evaluated to share the highest degree of similarity.

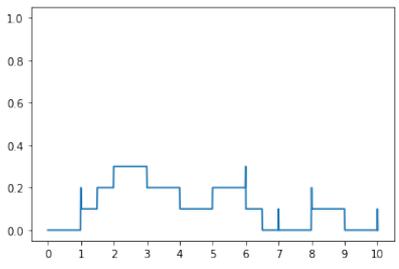

| Critic IAA Fuzzy Number | Similarity Measure ||||| 
|---|---|---|---|---|---|
| | Critic 1 | Critic 2 | Critic 3 | Critic 4 | Critic 5 |
| $A_1 = \{[1,1], [5,6], [2,3], [6,6], [1,4], [7,7], [8,9], [1.5, 6.5], [8,8], [10,10]\}$ | 1.0000 | 0.8635 | 0.8069 | 0.8925 | 0.9151 |
| $A_2 = \{[1,1], [6,7], [1,3], [6,10], [2,3], [8, 9.2], [9,10], [3,10], [8,8], [10,10]\}$ | 0.8635 | 1.0000 | 0.9344 | 0.9706 | 0.8802 |

| | | | | | | |
|---|---|---|---|---|---|---|
| | 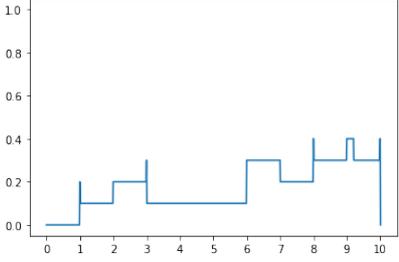 | | | | | |
| $A_3 = \{[1,1], [10,10], [4,7], [8,10], [7,8], [9,10], [9.5, 9.5], [1,10], [8,8], [10,10]\}$ | | | | | | |
| | 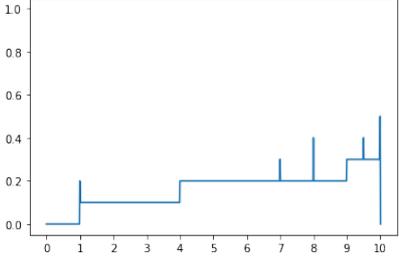 | 0.8069 | 0.9344 | 1.0000 | 0.9123 | 0.8231 |
| $A_4 = \{[1,1], [3,4], [1,3], [5,9], [3,3], [8,9], [9,10], [2, 9.2], [8,8], [10,10]\}$ | | | | | | |
| | 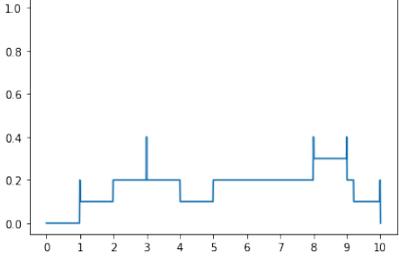 | 0.8925 | 0.9706 | 0.9123 | 1.0000 | 0.9088 |
| $A_5 = \{[1,1], [5,5], [4,5], [2,3], [2, 4.4], [9,10], [10,10], [8, 8.8], [8,8], [10,10]\}$ | | | | | | |
| | 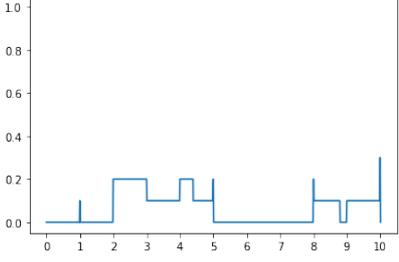 | 0.9151 | 0.8802 | 0.8231 | 0.9088 | 1.0000 |

**Fig. 5.** Similarity amongst the critic IAA fuzzy numbers.

## 7. Conclusion

As the IAA method is a relatively recent method for constructing aggregated fuzzy numbers, developing new measures for it requires the abstraction and application of contributions of studies involving other types of fuzzy numbers. In the case of this study, generalised fuzzy numbers were used as the primary source of inspiration. In this paper, a new method is proposed to measure the similarity between two IAA fuzzy numbers. To reach this goal, we define new features for similarity of IAA fuzzy numbers for the first time. In addition, we introduce a novel approach to construct the similarity measure based on these features using PCA as a machine learning technique. The proposed similarity measure was demonstrated utilising a diverse range of IAA fuzzy numbers, depicting adequate evaluation of little and substantial differences in structure. Without modification, the use of the proposed similarity is to provide a means of comparison between two IAA fuzzy numbers, and therefore is also means of comparison between sets of interval-valued data. We provided an illustrative example as a demonstration of the standard use of the proposed similarity measure, which was a comparison of films given interval scores, as well a comparison of film critics based on said scores. Similarly, the proposed similarity measure can be applied in different case studies to compare alternatives and decision-makers.

Potential future work would include updates to the proposed attributes, as base modifications have been made to their original definitions to allow for compatibility with IAA fuzzy numbers, though further appropriate modification would likely result in the better representation and separation of IAA fuzzy numbers. Similarity measures allow data types to be applicable in various areas of machine learning as a frequent requirement is the comparison between data points, therefore the contributions of this study is one such potential opening for the future practical application of aggregate fuzzy numbers from interval-valued data utilising the IAA method.


## References

[1] L. Yang, Q. Yang, Y. Li and Y. Feng, "K-Nearest Neighbor Model Based Short-Term Traffic Flow Prediction Method," in *18th International Symposium on Distributed Computing and Applications for Business Engineering and Science (DCABES)*, Wuhan, 2019.

[2] N. Altman, "An Introduction to Kernel and Nearest-Neighbor Nonparametric Regression," *The American Statistician,* vol. 46, no. 3, pp. 175-185, 1992.

[3] H. Yang, R. Jiang, C. Zhao and A. Li, "Evaluation of DDOS Attack Degree Based on GRA-TOPSIS Model," in *International Conference on Smart Grid and Electrical Automation (ICSGEA)*, Xiangtan, 2019.

[4] M. Collan and P. Luukka, "Evaluating R&D Projects as Investments by Using an Overall Ranking From Four New Fuzzy Similarity Measure-Based TOPSIS Variants," *IEEE Transactions on Fuzzy Systems,* vol. 22, pp. 505-515, 2014.

[5] C.-L. Hwang, Y.-J. Lai and T.-Y. Liu, "A new approach for multiple objective decision making," *Computers & Operations Research,* vol. 20, no. 8, pp. 889-899, 1993.

[6] S. Miller, C. Wagner and J. M. Garibaldi, "From Interval-Valued Data to General Type-2 Fuzzy Sets," *IEEE Transactions on Fuzzy Systems,* vol. 23, pp. 248-269, 2015.

[7] L. A. Zadeh, "Fuzzy Sets," *Information and Control,* vol. 8, no. 3, pp. 338-353, 1965.

[8] H. A. Khorshidi and S. Nikfalazar, "An improved similarity measure for generalized fuzzy numbers and its application to fuzzy risk analysis," *Applied Soft Computing,* vol. 52, pp. 478-486, 2016.

[9] K. Patra and S. K. Mondal, "Fuzzy risk analysis using area and height based similarity measure on generalized trapezoidal fuzzy numbers and its application," *Applied Soft Computing,* vol. 28, pp. 276-284, 2015.

[10] H. T. Xuan Chi and V. F. Yu, "Ranking generalized fuzzy numbers based on centroid and rank index," *Applied Soft Computing,* vol. 68, pp. 283-292, 2018.

[11] T. C. Havens, C. Wagner and D. T. Anderson, "Efficient Modeling and Representation of Agreement in Interval-Valued Data," *IEEE International Conference on Fuzzy Systems,* pp. 1-6, 2017.

[12] J. Navarro, C. Wagner, U. Aickelin, L. Green and R. Ashford, "Measuring Agreement on Linguistic Expressions in Medical Treatment Scenarios," *IEEE Symposium Series on Computational Intelligence (SSCI),* pp. 1-8, 2016.

[13] S. Miller, S. Appleby, J. Garibaldi and U. Aickelin, "Towards a More Systematic Approach to Secure Systems Design and Analysis," *International Journal of Software Engineering and Knowledge Engineering,* pp. 11-30, 2013.

[14] S. Nikfalazar, C.-H. Yeh, S. . E. Bedingfield and H. A. Khorshidi, "A hybrid missing data imputation method for constructing city mobility indices," in *Data Mining - 16th Australasian Conference*, Bathurst, 2018.

[15] D. Wu, J. M. Mendel and S. Coupland, "Enhanced Interval Approach for Encoding Words Into Interval Type-2 Fuzzy Sets and Its Convergence Analysis," in *International Conference on Fuzzy Systems*, Barcelona, Spain, 2010.



[16] F. Liu and J. M. Mendel, "Encoding Words Into Interval Type-2 Fuzzy Sets Using an Interval Approach," *IEEE Transactions on Fuzzy Systems,* vol. 16, no. 6, pp. 1503-1521, 2008.

[17] S. Miller, C. Wagner, J. M. Garibaldi and S. Appleby, "Constructing General Type-2 Fuzzy Sets from Interval-valued Data," *IEEE World Congress on Computational Intelligence,* pp. 1-8, 2012.

[18] S.-M. Chen, "New Methods for Subjective Mental Workload Assessment and Fuzzy Risk Analysis," *Cybernetics and Systems,* vol. 27, no. 5, pp. 449-472, 1996.


**Vitae**

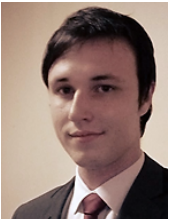

Justin Kane Gunn holds a Bachelor of Computer Science (BCS) from Monash University, and has recently completed his MSc thesis in the School of Computing and Information Systems at the University of Melbourne. Having worked as an Associate Lecturer at RMIT University, he currently works as a Lecturer and Course Coordinator at Stott's College whilst completing his MSc. His Research Interests include Machine Learning, Modelling, Uncertainty, Decision Making and Artificial Intelligence (Game Theory).

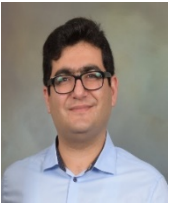

Hadi A. Khorshidi received his PhD from Monash University, where he worked as Research Fellow and Senior Data Analyst. He currently works as a Research Fellow in School of Computing and Information Systems at the University of Melbourne. His research areas include Decision Making and Optimization, Data Mining and Machine Learning, Modelling, Uncertainty and Digital Health, where he has published several papers.

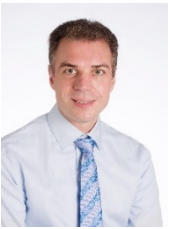

Uwe Aickelin holds a PhD degree from the University of Wales (UK). He is now a Professor and Head of School of Computing and Information Systems at the University of Melbourne. He is an Associate Editor of IEEE Transactions on Evolutionary Computation. His Research Interests include Artificial Intelligence (Modelling and Simulation), Data Mining and Machine Learning (Robustness and Uncertainty), Decision Support and Optimisation (Medicine and Digital Economy) and Health Informatics (Electronic Healthcare Records).